\documentclass[conference]{IEEEtran}
\IEEEoverridecommandlockouts
\usepackage{amsmath,amssymb,amsfonts}
\usepackage{algorithmic}
\usepackage{graphicx}
\usepackage{textcomp}
\usepackage{xcolor}
\usepackage{subfig}

\usepackage{booktabs}
\usepackage{multirow}

\usepackage[
     backend=biber,
     style=numeric-comp,
     sorting=none
     ]{biblatex} 
\begin{document}

\title{Hybrid Focal and Full-Range Attention Based Graph Transformers\\
}

\author{\IEEEauthorblockN{Minhong Zhu\textsuperscript{*}}
\IEEEauthorblockA{\textit{School of Biology and Basic Medical Science\thanks{ \textsuperscript{*} These two authors contributed equally.}} \\
\textit{Soochow University,}
Suzhou, China \\
}
\and

\IEEEauthorblockN{Zhenhao Zhao\textsuperscript{*},  Weiran Cai\textsuperscript{$\dagger$}}
\IEEEauthorblockA{\textit{School of Computer Science and Technology\thanks{ $\dagger$ Correspondence should be addressed to Weiran Cai: wrcai@suda.edu.cn.}\thanks{Codes are available at: https://github.com/minhongzhu/ffgt.}}\\
\textit{Soochow University,}
Suzhou, China \\
}

}

\maketitle

\begin{abstract}
The paradigm of Transformers using the self-attention mechanism has manifested its advantage in learning graph-structured data. Yet, Graph Transformers are capable of modeling full range dependencies but are often deficient in extracting information from locality. A common practice is to utilize Message Passing Neural Networks (MPNNs) as an auxiliary to capture local information, which however are still inadequate for comprehending substructures. In this paper, we present a purely attention-based architecture, namely \textbf{F}ocal and \textbf{F}ull-Range \textbf{G}raph \textbf{T}ransformer (FFGT), which can mitigate the loss of local information in learning global correlations. The core component of FFGT is a new mechanism of compound attention, which combines the conventional full-range attention with K-hop focal attention on ego-nets to aggregate both global and local information. Beyond the scope of canonical Transformers, the FFGT has the merit of being more substructure-aware. Our approach enhances the performance of existing Graph Transformers on various open datasets, while achieves compatible SOTA performance on several Long-Range Graph Benchmark (LRGB) datasets even with a vanilla transformer. We further examine influential factors on the optimal focal length of attention via introducing a novel synthetic dataset based on SBM-PATTERN. \textcolor{blue}{}
\end{abstract}

\section{Introduction}
Deep learning has shown its ability in modeling complex relationship in graph-structured data across diverse domains including chemistry, social science and biology. Facing the complex nature of data structure, the learning paradigm is under continuous evolution. The first surge of research interest was aroused by the message passing mechanism \cite{gilmer2017neural}, where messages are aggregated and propagated through neighbouring nodes to capture local structural information. This paradigm of Message Passing Neural Network (MPNN) has been extensively validated for structural and semantic representations \cite{kipf2016semi, xu2018powerful, velivckovic2017graph}, yet has also revealed its shortcomings known as over-smoothing \cite{li2018deeper} and over-squashing \cite{alon2020bottleneck, topping2021understanding}. Inspired by the success of Transformers in natural language processing \cite{vaswani2017attention} and computer vision \cite{dosovitskiy2020image}, the next generation of work has opened a new path in using self-attention mechanisms specifically for graph data, namely Graph Transformers (GTs). The key motivation is that by building long-range dependencies over the entire graph \cite{rampavsek2022recipe}, this mechanism is endowed with more expressive power through a graph-level scope and mitigate the deficiencies of MPNNs \cite{xu2018powerful}. 

Graph Transformers are expected to comprehend structures across scales and the underlying semantic information. However, the lack of strong inductive bias \cite{dosovitskiy2020image} often makes the canonical Transformers struggle when applied to graph data. An effective way of retrieving structural information is to use positional encoding \cite{muller2023attending}. Models following this paradigm include leveraging absolute positional encoding \cite{kreuzer2021rethinking, dwivedi2020generalization}, pair-distance information \cite{ying2021transformers, park2022grpe}, graph kernels \cite{mialon2021graphit}. \cite{zhang2022autogt} summarizes a unified framework of this kind where positional encoding can be selectively added to one or multiple locations within input representations, attention and values. Notably, a prominent feature with GTs is the rich possibilities of incorporating edge information into the inductive bias \cite{ying2021transformers, park2022grpe, chen2023graph, ma2023graph}, which often helps to achieve SOTA performance \cite{chen2023graph, ma2023graph}.

However, GT models, proficient in establishing global dependencies, are still inadequate in the perception of local substructures that are often intrinsic to specific graph categories. An effective compound approach is to compensate GTs by utilizing modules to explicitly enhance information aggregation from locality. A common practice is utilizing MPNNs as an auxiliary to the global self-attention \cite{min2022transformer}. Existing methods include building Transformer blocks on top of MPNNs \cite{wu2021representing, rong2020self}, alternately stacking MPNN blocks and Transformer blocks \cite{lin2021mesh}, and parallelizing MPNN blocks and Transformer blocks \cite{rampavsek2022recipe, zhang2020graph}. A typical example is GraphGPS \cite{rampavsek2022recipe} combining Transformer layers (with positional encoding) and MPNNs into a unified framework, which achieves SOTA performance. Yet, there still exist several flaws by utilizing MPNNs: (\romannumeral1) Using stacked MPNN modules is known to suffer from over-squashing and will lead to information loss. (\romannumeral2) For substructures with a larger diameter ($\geq 2$), applying MPNN layer-wise will compromise the integrity of substructural information.
(\romannumeral3) Relying solely on MPNNs to process edge information is insufficient in extracting the relationship among nodes and edges \cite{chen2023graph}.

In this work, we propose a purely attention-based architecture, the \textbf{F}ocal and \textbf{F}ull-Range \textbf{G}raph \textbf{T}ransformer (FFGT), which is proficient in acquiring both global and local information. A typical FFGT layer realizes an attention mechanism at different scales through two compensating modules: While the full-range attention module follows the canonical design of attention layer that acquire global correlations, it is compensated by a focal attention module that focuses on local ego-nets with a properly chosen receptive field. By hybridizing attention at different scales, the interaction between intrinsic local substructures and graph-level information can be effectively captured. While this compound mechanism enhances existing global attention-based GTs with the desirable substructure-awareness, it out-competes models based on limited receptive fields where access to graph-level information is not rendered, such as GT-Sparse \cite{dwivedi2020generalization} (only nearest neighbour nodes correlated), K-hop MPNN \cite{nikolentzos2020k} (no acquisition of global information) and SAN \cite{kreuzer2021rethinking} (low attention on distant nodes). Another crucial advantage of the proposed architecture is in the improved acquisition of local substructural information. Compared to convolution-based models such as K-hop MPNNs (will be discussed in Section \ref{method}), the focal attention module provides far more flexibility allowing exploration in a broader solution space. With the above mentioned two merits, the proposed framework can benefit us by correlating information at different scales through the compound attention mechanism, which may serve as an important perspective towards utilizing graph transformers.

The rest of the paper will include the following parts:
\begin{itemize}
    \item We propose FFGT, a purely attention-based architecture that hybridizes the local substructural information of ego-nets and the global information over the entire graph. The architecture can be flexibly embodied by advanced attention techniques without increasing the complexity.
    \item We carry out extensive experiments on real-life benchmarks containing small-sized to large-sized graphs to evaluate the effectiveness of our approach. Even with simple attention modules adapted from a vanilla Transformer \cite{vaswani2017attention}, our model can match or outperform SOTA methods on datasets from Long Range Graph Benchmark (LRGB) \cite{dwivedi2022long}
    \item We identify the importance of the intrinsic scale of substructures on different graphs. We design a new type of tests based on synthetic datasets generated with the Stochastic Block Model (SBM) \cite{dwivedi2020benchmarking} to illustrate how our model is able to identify such variation of substructures through the change of focal length.
\end{itemize}

\section{Related Work}

\subsection{Graph Transformers with Improved Attention} 
The success of Transformers in NLP and CV has spurred recent endeavors to craft specialized Transformers tailored for graph data. However, integrating structural information into the canonical Transformer architecture is not a straightforward task, owing to the intricate nature of graph structures. An effective strategy is to enhance attention computation by introducing supplementary structural information.
Graphormer\cite{ying2021transformers} adds spatial encoding and edge encoding to attention matrix to provide extra structural information. Such idea is further improved in GRPE \cite{park2022grpe} by introducing node-spatial and node-edge relations. 
In the recent work CSA \cite{menegaux2023self}, the attention module is enriched by Chromatic Self-Attention (CSA), which selectively modulates message channels between two nodes. Different from its predecessors, GRIT \cite{ma2023graph} employs virtual edge representations to compute the attention matrix representation. By iteratively updating the edge representation layer by layer, GRIT effectively amalgamates structural information with real edge information.
Besides, some previous works treat both nodes and edges as tokens \cite{hussain2022global, kim2022pure} and theoretically result in the expressiveness boost.

\subsection{MPNN-Auxiliary Graph Transformers} 
Mounting MPNNs is another popular strategy to involve structural information in GTs.
GraphTrans\cite{chen2023graph} adds Transformer blocks after standard GNN blocks, with the intention to enhance the local awareness gained from MPNNs. 
Mesh Graphormer \cite{lin2021mesh} stacks a Graph Residual Block (GRB) on
a Transformer layer to model both local and global interactions among 3D mesh vertices and body joints.
The recent work GraphGPS \cite{rampavsek2022recipe} further improves the idea by providing a unified recipe for designing GTs. It integrates the MPNN-based models with global attention (with extra PE/SE) to enhance the model's expressivity.

\subsection{Subgraph GNNs} 
Subgraph GNNs apply MPNNs to a set of subgraphs extracted from the input graph and aggregate the resulted representations in order to break the expressivity bottleneck of MPNNs. SubGNN \cite{alsentzer2020subgraph} employs a message-passing framework to propagate messages from a set of anchor patches to the subgraph, enabling the capture of distinctive properties within the subgraph. SUN \cite{frasca2022understanding} unifies Subgraph GNN architectures and proves that they are bounded by 3-WL test. Another thread of work focuses on developing K-hop MPNNs \cite{nikolentzos2020k} and promotes the expressivity by injecting subtructural information \cite{feng2022powerful}. Following the advances in subgraph-involving GNNs, the concept of subgraph-level modeling is integrated into Graph Transformers. Graph ViT/MLP-Mixer \cite{he2023generalization} leverages the ViT/MLP Mixer architecture introduced for CV by tokenizing subgraphs to achieve linear time/memory complexity. Similar ideas are adopted in work that attempts to scale GTs \cite{chen2022nagphormer, shirzad2023exphormer,  ngo2023multiresolution}.

\begin{figure}[t]
\centering  
{

\includegraphics[width=0.8\linewidth,height=8cm]{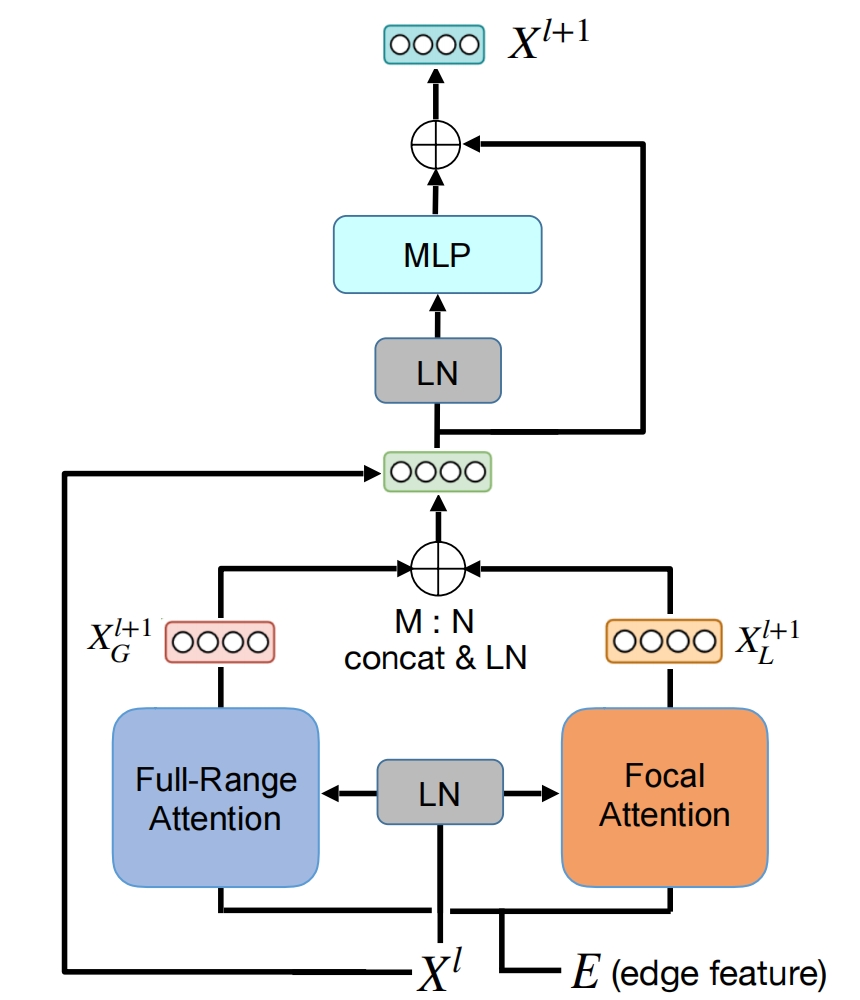}}

\setlength{\abovecaptionskip}{1.cm}
\centering
\caption{FFGT layer of compound attention consisted of compensating scopes: the full-range attention block is a fully-connected attention block for obtaining global correlations, while the focal attention block is an attention module delimited to local ego-nets for comprehending substructural information. Each attention block is comprised of a number of attention heads (M and N respectively). The output of each layer is concatenated and normalized (Layer Normalization here) to allow integration from different scopes. Edge features are used by the two attention blocks separately.
}
\label{fig.FFGT}
\end{figure}

\section{Methods}
\label{method}

We propose a framework that integrates full-range attention over the entire graph and focal attention on local ego-nets. This framework has the merits of merging different scopes of attention, encapsulating much richer information of local substructures and mitigating over-smoothing and over-squashing brought by MPNNs. Moreover, the purely attention-based framework enables various ways of employing and updating edge information.

\begin{figure}
    \begin{minipage}[b]{0.40\columnwidth}
	\subfloat[][]{\hspace{-0.3cm}\includegraphics[width=1.5\linewidth,height=3.5cm]{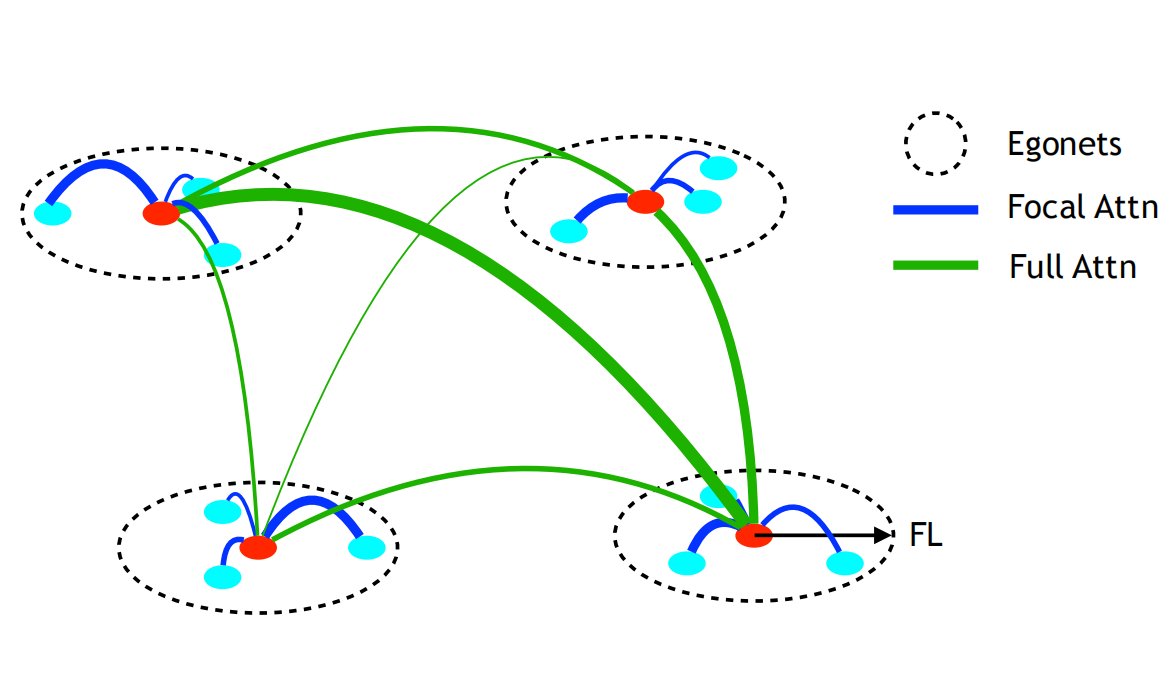} \label{fig.egonet_stru}}
	\subfloat[][]{\includegraphics[width=1.2\linewidth,height=3.9cm]{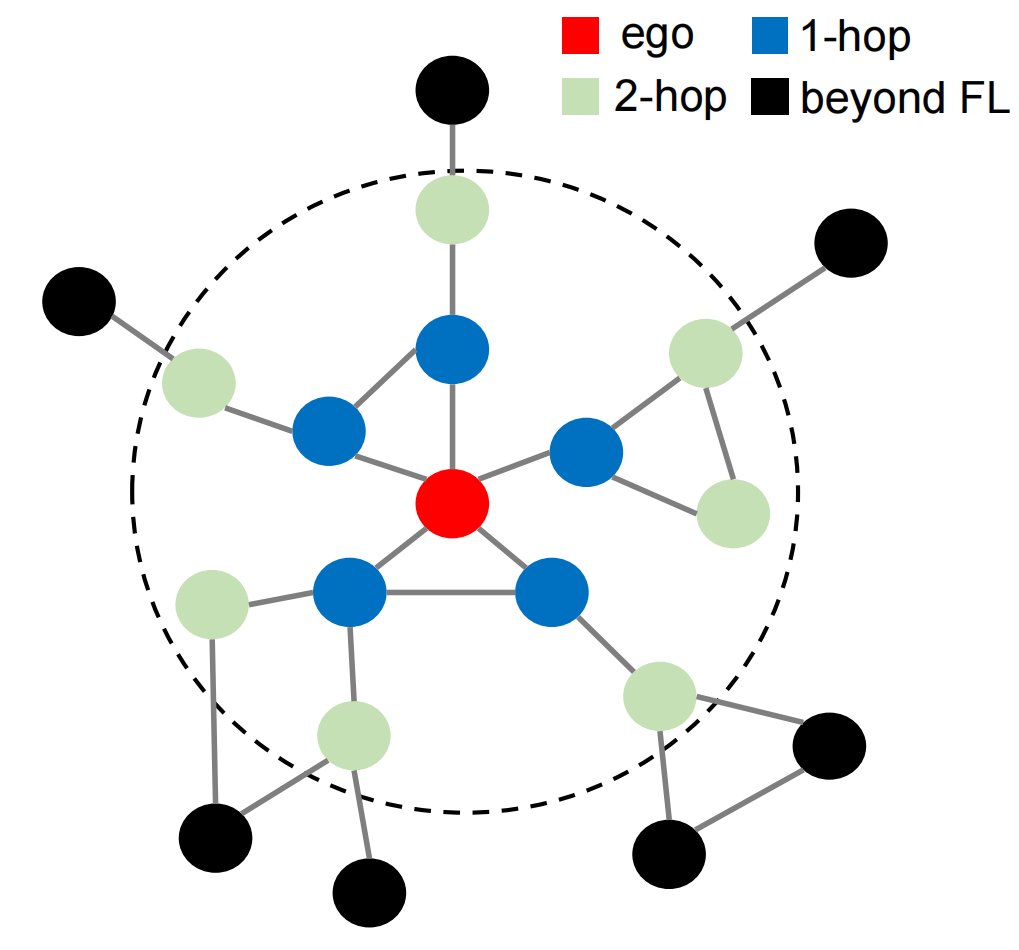} \label{fig.egonet_local}}
 \quad
    
    \subfloat[][]{\hspace{1.5cm}\vspace{-1.8cm}\includegraphics[width=2\linewidth,height=8.5cm]{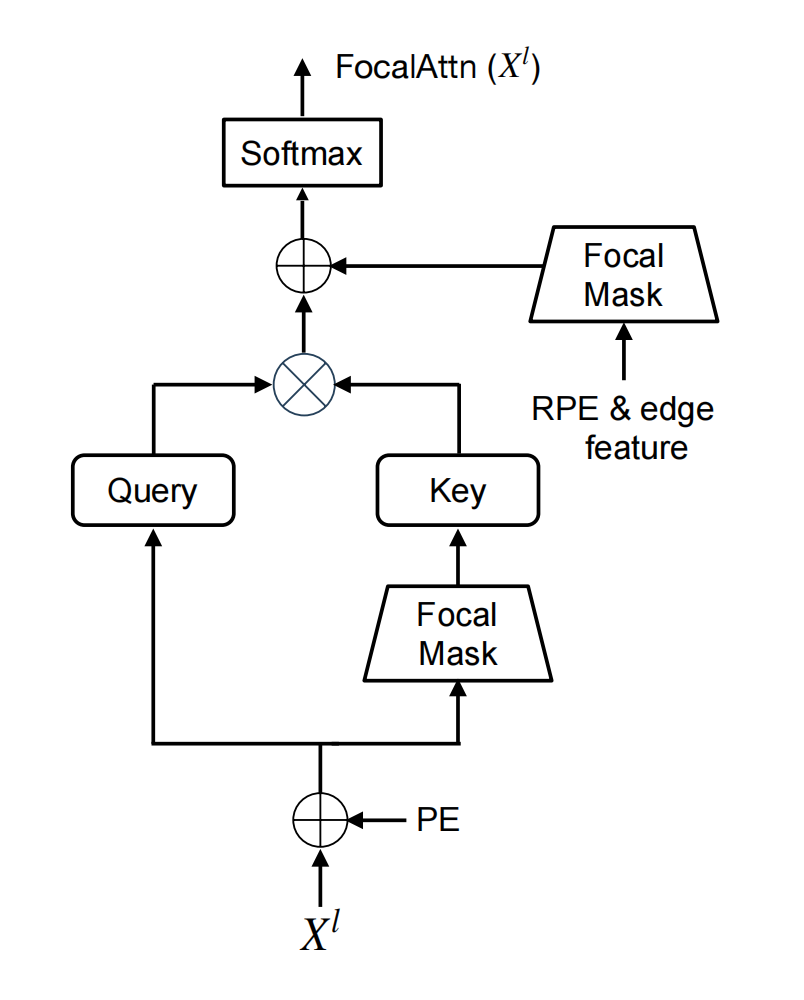} \label{fig.Hopmask}} 
\end{minipage}
\setlength{\abovecaptionskip}{0.5cm}
	\caption{Focal Attention Block. \textbf{(a)} Focal Attention prunes high correlations over long ranges (green lines), allocating more focus to the locality (blue lines); \textbf{(b)} Example of a local ego-net with $FL=2$. Nodes (black) outside the circle (i.e. with distance greater than 2 hops) are excluded from attention computation; \textbf{(c)} Detailed architecture inside the focal attention block. Focal Mask $FM \in \mathbb{R} ^{n\times n}$ is employed to exclude nodes beyond the ego-net range, with $FM_{ij} = 1$ only when node $j$ belongs to the K-hop ego-net centered on node $i$ and $FM_{ij} = 0$ otherwise. }
\end{figure}

\subsection{FFGT layer}
Here we introduce the FFGT layer of compound attention. We first define the paradigm that can be instantiated with a general attention functionality and give a specific instantiation of the model in the next.

\subsubsection{Paradigm of Compound Attention} A standard FFGT layer consists of a full-range attention layer to capture global information and a focal attention layer to extract local substructural information. The full-range attention layer can be any unbounded fully connected attention layer, whereas the focal attention layer is allowed to employ another attention mechanism that is more proper for perceiving local substructures. Likewise, the ratio of the number of attention heads $M:N$ of both modules can be optimally set. The interaction of the two modules is rendered through aggregating the output of all attention heads. The architecture is illustrated in Fig. \ref{fig.FFGT} (residual connections with layer normalization are omitted for clarity), which is formulated below

\begin{equation}\label{aggr}
X^{l+1} = MLP(Aggr(X_{G}^{l+1}, X_{L}^{l+1}))
\end{equation}
\begin{equation}\label{full}
X_G^{l+1} = FullAttn(X^l)
\end{equation}
\begin{equation}\label{focal}
X_L^{l+1} = FocalAttn(X^l)
\end{equation}

\noindent where $FullAttn$ and $FocalAttn$ are instances of a global attention mechanism and a delimited attention mechanism with learnable parameters, respectively. $Aggr$ is the modular aggregation strategy, which can be realized by a concatenation function of all heads of both attention modules, followed by a linear layer further merging them; $MLP$ is a 2-layered MLP block.

\subsubsection{Focal Attention} To establish auxiliary local information representation of subgraphs, we deploy multiple heads for attention of a delimited range in a GT. A K-hop ego-net centered at each node is extracted on which the focal attention with a local inductive bias is used to explore the nodal correlations (Fig. \ref{fig.egonet_stru}). The node representation can be updated as follows

\begin{equation}\label{focal2}
X_i = \sum_{j\in Ego\{i, FL=k\}}Attn(i,j|local~bias)V_j
\end{equation}

\noindent where $Ego\{i, FL=k\}$ refers to the K-hop ego-net rooted on node $i$; $Attn(i,j)$ is the attention score between node $i$ and node $j$; $V_j$ is the value of node $j$. The scale of the ego-net is defined by the focal length ($FL$), a hyperparameter critically related to the intrinsic substructure of the studied graphs.

The superiority of focal attention in capturing local information can be expected with the comparison to K-hop MPNNs \cite{nikolentzos2020k, yao2023improving}. The commonality between the two lies in the fact that with both, the root node receives messages from all nodes within its K-hop ego-net. However, with the K-hop MPNN, the same learnable weights are shared in aggregating local information at the same distance. In contrast, each focal attention head can take advantage of the nodes' relative distances as a bias and learns a pattern for aggregating message from all nodes within the ego-net. The latter enjoys full flexibility in treating the local network, benefited from relaxing largely the limitation of MPNNs in the aggregation, which is bounded by the requirement of invariance in the aggregating order. This means that only when the same attention weights are allocated to the nodes at the same distance, the functionality of focal attention degenerates to a K-hop MPNN layer. Therefore, the focal attention, by exploring a greater space of solutions, will be more powerful than K-hop MPNNs, which promises its ability in capturing local information.

\subsection{Instantiation}
\label{section:Instantiation}
The FFGT is highly flexible allowing independent selection of the attention mechanism used for attention modules of different scopes. Despite the choices, we demonstrate here how to embed the attention mechanisms in our architecture that are most commonly used in GTs, following the paradigm proposed in AutoGT \cite{zhang2022autogt}. To be specific, we compute the attention map as follows

\begin{equation}\label{qk}
Q = XW_Q,\hspace{1em} K=XW_K
\end{equation}

\begin{equation}\label{Attn}
Attn(X) = softmax(\frac{QK^T}{\sqrt{d}}+B)\odot C
\end{equation}

\noindent where $W_Q\in \mathbb{R} ^{d\times d}$ and $W_K\in \mathbb{R} ^{d\times d}$ are learnable parameters for queries and keys. Matrix $B\in \mathbb{R} ^{n\times n}$ is the attention bias generally composed of the combination of relative positional encoding (RPE) \cite{park2022grpe, ying2021transformers, zhao2021gophormer} and edge features \cite{park2022grpe, ying2021transformers}. $C\in \mathbb{R} ^{n\times n}$ is an optional gate to enhance the relative positional encoding \cite{luo2022your}. For those requiring absolute positional encoding, the node input can be augmented by positional encoding (PE) prior to entering the attention layer. 

The delimited attention in focal attention layer is achieved through the $FocalMask$ function applied to the keys, RPE and edge features, as shown in Fig. \ref{fig.Hopmask}. $FocalMask$ is a binary matrix consisting of 0 and 1, which excludes nodes beyond the ego-net range from the attention computation. In comparison to using an Attention Mask \cite{min2022masked}, the advantage of employing $FocalMask$ lies in its omission of correlation calculations involving the center node and nodes outside its ego-net. This enables the reduction of computation from quadratic complexity to a constant complexity, and thus does not alter the scalability. Formally, the focal attention module can be expressed as follows

\begin{equation}\label{Attn2}
Attn(X) = softmax\left(\frac{Q\times FM(K)^T}{\sqrt{d}} + FM(B)\right) \odot C
\end{equation}

\noindent where $FM$ is the Focal Mask function, which is a $\mathbb{R}^{n\times n}$ matrix with $FM_{ij} = 1$ if $j\in Ego(i,FL=K)$ and $FM_{ij} = 0$ otherwise.

\begin{table*}[]
\setlength{\abovecaptionskip}{0.1cm}
\setlength{\belowcaptionskip}{-0.0cm}
\caption{Comparison of FFGT with fully-connected GTs. Results are averaged over 4 runs.}
\centering
\renewcommand\arraystretch{1.5}
\begin{tabular}{@{}lcccc@{}}
\toprule
\multirow{2.5}{*}{\quad Model} & ZINC    & Ogbg-Molpcba  & Peptide-Func   & Peptide-Struct  \\ 
\cmidrule(l){2-5} 
\multicolumn{1}{c}{}    & MAE↓   & AP$\uparrow$    & AP$\uparrow$    & MAE↓  \\ 
\midrule
\quad VANILLA     & 0.156±0.006   & 0.2638±0.0058  & 0.6648±0.0087     & 0.2500±0.0027 \\
\quad VANILLA-FFGT     & \textbf{0.140±0.005}      & \textbf{0.2779±0.0049}     & \textbf{0.6914±0.0040}    & \textbf{0.2456±0.0017}    \\ 
\midrule
\quad GRPE    & 0.101±0.006   & 0.2856±0.0039      & 0.6836±0.0060     & 0.2436±0.0018     \\
\quad GRPE-FFGT    & \textbf{0.093±0.004}  & \textbf{0.2902±0.0010}     & \textbf{0.6956±0.0080}    & \textbf{0.2416±0.0026} \\ 
\bottomrule
\end{tabular}
\label{table.FFGT}

\end{table*}

\section{Experiments}
\subsection{Implementation Details}
\subsubsection{FFGT Attention Module} 
Considering the composition of FFGT attention modules, one simple choice is to let the focal layer share the same correlation attention mechanism of full-range but within the focal range, although the choice of a distinct function may meet further needs. Here, we use this simplification and choose mechanisms used in Vanilla Transformer and GRPE to construct Vanilla-FFGT and GRPE-FFGT, respectively. By incorporating Vanilla Transformer, we are interested in testing how well the hybridization of Focal and Full-Range attention alone can benefit the learning, in spite of many alternatives. We also choose the more sophisticated GRPE because it involves a combination of RPE and Edge Encoding as attention bias, which can serve to examine our framework when effective inductive biases are introduced.

In the Vanilla-FFGT, we follow the Transformer+LapPE architecture poposed in \cite{dwivedi2022long}. Since the architecture in \cite{dwivedi2022long} does not involve edge information, we add edge encoding as attention bias for fair comparison. In GRPE-FFGT, we directly utilize the attention mechanism with edge features as used in the original. While the number of attention heads is not required to match between the full-range and focal attention modules, we have chosen the ratio to be 1. This is a prior since we cannot determine whether global or local information holds the greater significance.

\subsubsection{Virtual Node} 
For graph-level prediction tasks, following Graphormer, we adopt a special node called virtual node. It is connected to all real nodes in both full-Range and focal attention modules, which acts as a classification token \cite{devlin2018bert}.

\subsection{Results on Graph Benchmarks}
We primarily focus on evaluating the performance of the FFGT on graph-level tasks, where both local and global information is generally influential. Two small molecule benchmarks and two benchmarks from Long Range Graph Benchmarks are chosen. The former pair of benchmarks centers on small graphs where local information plays a pivotal role, while the latter pair is tailored for gauging a model's ability of learning long-range dependencies. First, we compare FFGT with their backbone models separately on the four benchmarks. Then, we compare Vanilla-FFGT and GRPE-FFGT with the baselines containing SOTA Graph Transformer models and other GNN-based models. For completeness, we further conduct extra evaluations on node-level tasks to demonstrate the effectiveness of FFGT in capturing local information in various graph types.

\subsubsection{Molecule Property Prediction} We first evaluate FFGT on two small molecule datasets, ZINC from Benchmarking GNNs \cite{dwivedi2020benchmarking} and Ogbg-Molpcba from Open Graph Benchmark (OGB) \cite{hu2020open}. For ZINC, we use Mean Absolute Error (MAE) as an evaluation metric for the regression task. For Ogbg-Molpcba, as the class balance is skewed (only 1.4\% of data is positive) and the datasets contain multiple classification tasks \cite{hu2020open}, we use the Average Precision (AP) averaged over the tasks as the evaluation metric. As shown in Table \ref{table.FFGT}, FFGT promotes the performance on both small molecule benchmarks. The results demonstrate that for small molecule datasets where local information plays an important role, it is beneficial to employ global attention in alliance with focal attention as to mitigate the shortcomings of the former in understanding local substructures. On the ZINC dataset, we observe that the FFGT achieves the best performance when the focal length is set to 1, which aligns with the advantage of integrating a local MPNN with global attention as illustrated in GraphGPS \cite{rampavsek2022recipe}. In comparison, interestingly, despite the similar average graph size, the optimal focal length for Ogbg-Molpcba turns out to be 3. Such offset indicates that considering subgraphs beyond 1-hop neighbourhood is beneficial for identifying the heterogeneity among graphs due to the detailed substructures.


\subsubsection{Long Range Graph Benchmark (LRGB)} We further evaluate the FFGT architecture on Peptide-Functional and Peptide-Structural datasets from LRGB \cite{dwivedi2022long}. They are intended to test the model's ability to capture long-range dependencies in the graph. Similarly, we use MAE in the regression task on Peptide-Structural and AP in the imbalanced multi-task classification on Peptide-Functional for evaluation. As shown in Table \ref{table.FFGT}, FFGT enhances the performance of the two backbone models on both datasets with a statistical significance. A remark is that the backbone model GRPE with its relative positional encoding has already sufficed to characterise the relations among atoms associated with the Peptides-Structural dataset and thus leaves little space to its variant with FFGT for further improvements. 

\begin{table*}[]
\setlength{\abovecaptionskip}{0.1cm}
\setlength{\belowcaptionskip}{-0.0cm}
\caption{Comparison of FFGT with SOTA GTs and GNN-based models. We use "-" for missing values from literatures, bold for the best performance, and underlines for those where FFGT are compatible with the best models. Results are averaged over 4 runs.}
\centering
\renewcommand\arraystretch{1.5}
\begin{tabular}{@{}lcccc@{}}
\toprule
\multirow{2.5}{*}{\quad Model}  & ZINC    & Ogbg-Molpcba  & Peptide-Func   & Peptide-Struct  \\ 
\cmidrule(l){2-5} 
\multicolumn{1}{c}{}    & MAE↓   & AP$\uparrow$   & AP$\uparrow$    & MAE↓  \\ 
\midrule
\quad GCN     & 0.367±0.011   & 0.2424±0.0034      & 0.5930±0.0023     & 0.3234±0.0006 \\
\quad GatedGCN     & 0.226±0.014   & 0.2670±0.0020      & 0.5864±0.0077     & 0.3420±0.0013 \\
\quad SUN    & 0.084±0.002   & -      & 0.6730±0.0078     & 0.2498±0.0008     \\
\midrule
\quad SAN + LapPE    & 0.139±0.006      & 0.2765±0.0042     & 0.6384±0.0121    & 0.2683±0.0042    \\
\quad GPS    & \textbf{0.070±0.004}      & \textbf{0.2907±0.0028}     & 0.6535±0.0041    & 0.2500±0.0012    \\
\quad Exphormer    & -     & -     & 0.6527±0.0043    & 0.2481±0.0007    \\
\quad MGT    & 0.131±0.003      & -     & 0.6817±0.0064    & 0.2453±0.0025    \\
\quad Graph ViT    & 0.085±0.005  & -    & 0.6942±0.0075    & 0.2449±0.0016  \\ 
\midrule
\quad VANILLA-FFGT     & 0.140±0.005  & 0.2779±0.0049    & \underline{0.6914±0.0040}    & \underline{0.2456±0.0017}  \\
\quad GRPE-FFGT     & 0.093±0.004  & \underline{0.2902±0.0010}    & \textbf{0.6956±0.0080}    & \textbf{0.2416±0.0026}  \\
\bottomrule
\end{tabular}
\label{table.comp}

\end{table*}

\subsubsection{Model Comparison} We compare the FFGT with the state-of-the-art GTs and GNN-based models.
 The selection of baselines is intended to examine the effectiveness of FFGT. We use MPNNs and Subgraph GNNs (GCNs \cite{kipf2016semi}, GatedGCN \cite{bresson2017residual}, SUN \cite{frasca2022understanding}) to show the necessity of global information. SAN \cite{kreuzer2021rethinking} is a Graph Transformer largely limited to the neighbourhood, but puts emphases on the role of full-range attention as well. GraphGPS \cite{rampavsek2022recipe} and Exphormer \cite{shirzad2023exphormer} hybrid Transformers and MPNNs, which are suitable to demonstrate when focal attention would be needed. Finally, Graph ViT \cite{he2023generalization} and MGT \cite{ngo2023multiresolution} implements coarsening-related strategies to learn subgraph-level representations, which contrasts with ours in acquiring information from substructures. The results are shown in Table \ref{table.comp}. We observe that the FFGT models surpass baselines significantly on two benchmarks from LRGB. 

 Yet, although witnessing significant improvement compared to the backbone model, GRPE with FFGT does not perform better on ZINC and Ogbg-Molpcba or is even inferior compared to some of the baselines. Such a gap generally comes from the mismatch of backbone model and datasets. Since the finest-grained local substructures are crucial in the prediction on these datasets, the purely locality-based MPNN+PE model is already able to achieve an excellent performance \cite{rampavsek2022recipe}, whereas a Transformer module plays a limited role. With GRPE or the Vanilla Transformer (without FFGT) largely underperforming on the two datasets, it is hard to surpass the SOTA models by simply using the FFGT, even with a significant boost achieved by this means. Using a stronger backbone models with the aid of FFGT may generate a compatible or even surpassing result. 

\subsubsection{Node Classification} For completeness, we test the FFGT architecture on node classification tasks to further illustrate its ability in comprehending local information. We use homophily Citation Networks (Cora, Citeseer \cite{sen2008collective}) and heterophily Wikipedia Network (Chameleon \cite{rozemberczki2021multi}) for evaluation. We compare the performance with the Graph Transformers originally designed for graph-level tasks (SAN, Graphormer), and with those designed specifically for node-classification tasks (UniMP \cite{shi2020masked}, NaGphormer \cite{chen2022nagphormer}).

The results are shown in Table \ref{table.node}, where the accuracy is used as evaluation metric since the datasets are generally balanced. Although not designed to tackle node-level tasks, FFGT improves the performance of GRPE on all datasets significantly, which demonstrates its ability to enhance local information capturing on various graph data. Without additional modifications in the attention mechanism, GRPE-FFGT can already out-perform UniMP and NAGphormer, which were specifically designed for node classification tasks. It is yet fair to note that MPNNs usually performs better on these node-level tasks than Graph Transformers.

\begin{table}[]
\setlength{\abovecaptionskip}{0.1cm}
\setlength{\belowcaptionskip}{-0.0cm}
\caption{Evaluation of FFGT on node-level tasks. We use "-" for the missing values from literature, bold for the best performance.}
\centering
\tabcolsep=0.45cm
\renewcommand\arraystretch{1.5}
\begin{tabular*}{\linewidth}{@{}lccc@{}}
\toprule
\multirow{2.5}{*}{\quad Model}  & Cora    & CiteSeer  & Chameleon   \\ 
\cmidrule(l){2-4} 
\multicolumn{1}{c}{}    & Acc.(\%)$\uparrow$  & Acc.(\%)$\uparrow$   & Acc.(\%)$\uparrow$    \\ 
\midrule
\quad SAN    & 81.91±3.42     & 69.63±3.76     & 55.62±0.43      \\
\quad Graphormer    & 67.71±0.78     & 73.30±1.48      & 36.81±1.69        \\
\quad UniMP    & 84.18±1.39     & 75.00±1.59      & -       \\
\quad NAGphormer    & 85.77±1.35      & 73.69±1.48    & -      \\
\midrule
\quad GRPE     & 82.52±1.30  & 72.59±1.40    & 54.65±1.50     \\
\quad GRPE-FFGT     & \textbf{86.59±0.83}  & \textbf{75.31±1.32}    & \textbf{62.52±0.97}   \\
\quad $\Delta$ \textrm{Acc-FFGT}   & + 4.07  & + 2.72    & + 7.87   \\
\bottomrule
\end{tabular*}
\label{table.node}

\end{table}

\subsection{Ablation Study and Substructural Awareness}

Substructures play an important role in real-world graph data, such as the functional groups in small molecules, the amino acid residues in peptides, and the communities in social networks. One major difficulty in learning these substructures is that their scales can vary greatly. There is usually characteristic scale that is referred to the typical size of fundamental substructures within which nodes are highly correlated with each other (in terms of for example the average shortest path distance). In the following section, we explore the optimal choice of focal length in different types of graphs, and illustrate that compared to backbone models, FFGT can effectively capture the different characteristic scales of the key substructures within them.

\subsubsection{Tests on Empirical Datasets}
\label{Empirical Datasets}
We evaluate the effect of different choices of $FL$ on two empirical datasets: ZINC and Peptide-functional. For small molecules in ZINC, important functional groups usually contains only 3-5 atoms (e.g. carboxyl and aldehyde groups), with multivariate ring (e.g. benzene ring) as an exception. Conversely, the key substructures in peptides in the Peptide-Functional dataset are amino acid residues, which can have up to 14 atoms in their side chains. This indicates that the scales of the substructures appearing in the two datasets differ significantly.

\begin{table*}[]
\setlength{\abovecaptionskip}{0.1cm}
\setlength{\belowcaptionskip}{-0.0cm}
\caption{Performance of Vanilla-FFGT on different SBM-PATTERN datasets as $FL$ varies. Results are averaged over 8 runs.}
\centering
\renewcommand\arraystretch{1.5}
\begin{tabular}{@{}lcccc@{}}
\toprule
\multicolumn{1}{c}{Model} & p=0.16    & p=0.14  & p=0.12   & p=0.10  \\ 
\midrule
\quad Vanilla     & 87.3625±0.0699   & 84.2871±0.0729     & 81.3535±0.0650     & 78.9624±0.0378 \\
\midrule
\quad $FL=1$     & \textbf{87.6101±0.0995} & \textbf{84.4141±0.0406}   & 81.4364±0.0634     & 78.9674±0.0464 \\
\quad $FL=2$     & 87.4651±0.0843    & 84.3915±0.0688      & \textbf{81.4489±0.0426}    & \textbf{79.0464±0.0435}    \\ 
\quad $FL=3$     & 87.4412±0.0709   & 84.3270±0.0567      & 81.4277±0.0745    & 79.0292±0.0564    \\
\bottomrule
\end{tabular}
\label{table.pattern}
\end{table*}

\begin{figure*}[t]   
  \centering           
  \subfloat[ZINC]   
  {
      \label{fig.zinc}\includegraphics[width=0.45\textwidth]{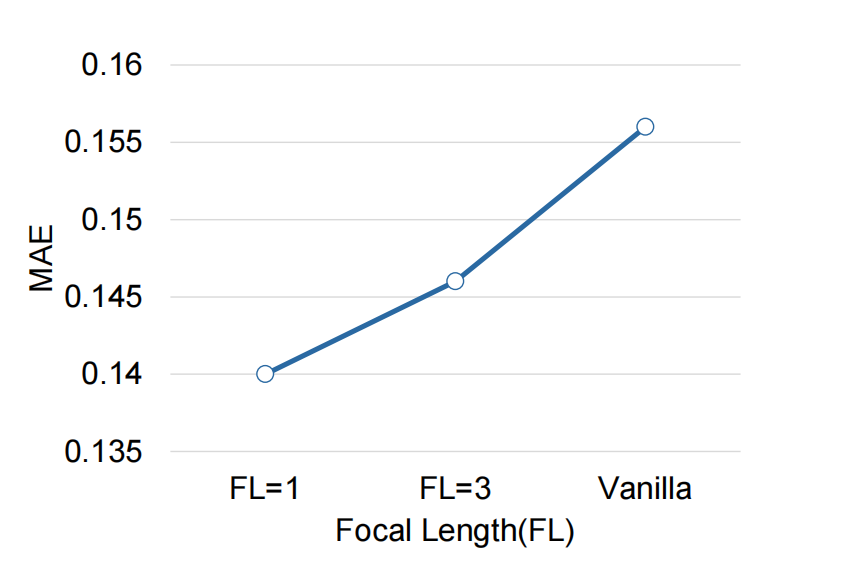}
     
  }
  \quad
  \subfloat[Peptide-Functional]
  {
      \label{fig.pepfuc}\includegraphics[width=0.45\textwidth]{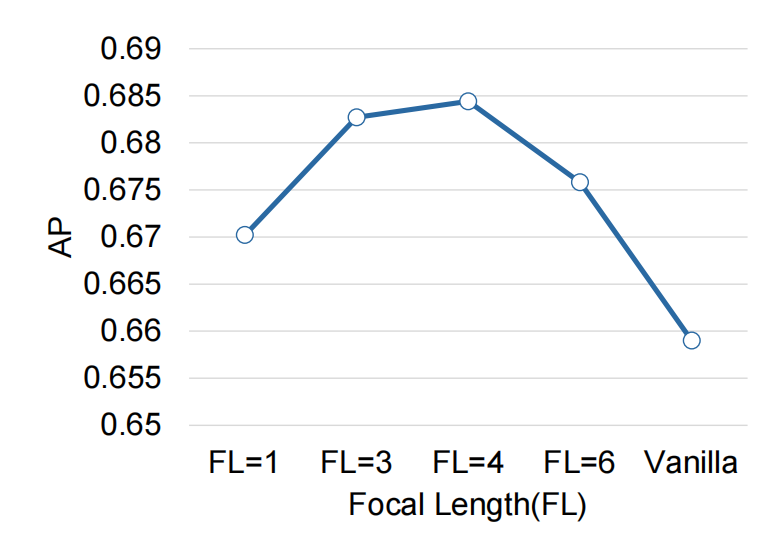}
  }
  \setlength{\abovecaptionskip}{0.2cm}
\caption{Ablation study with the focal length ($FL$) on ZINC and Peptide-Functional datasets. Vanilla-FFGT is used here as a backbone model. The horizontal axis represents the focal length ($FL$) and ``Vanilla" refers to the fully-connected version.}     
  \label{Fig.Ablation}            
\end{figure*}

The results of the ablation tests indeed reveal the characteristic scales, as shown in Figure \ref {Fig.Ablation}. (We refer to $FL = 0$ as the Vanilla, which represents the raw backbone model). For ZINC, the best performance is reached at $FL = 1$; in contrast, for Peptide-Functional, the performance peaks at $FL = 4$. It is especially enlightening that the maximum number of carbon atoms in the side chain of amino acids is 4, which means that it takes 4 hops for alpha carbon atoms to "see" all the main atoms of the side chains. This indicates that focal attention can learn such a characteristic of amino acids, while a decreasing performance for $FL>4$ indicates the lack of recognition of such key substructures with other attention scopes. As for ZINC, $FL=1$ is already sufficient for central atoms in most functional groups to "see" each other, which matches its determinants of extreme locality. The above results hence show that the optimal focal length in FFGT aligns with the right scale of the substructure on the graph.

\subsubsection{Controlled Tests on SBM-PATTERN}

To further explore the model's ability to distinguish such characteristic scales of substructure, we design a synthetic graph dataset with controllable parameters to examine the dependence of the optimal $FL$ explicitly. More concretely, we adapt SBM-PATTERN based on the Stochastic Block Model (SBM), a synthetic dataset focusing on inductive node prediction, and generate graphs with communities of tunable scales.  

\subsubsection*{SBM-PATTERN} PATTERN is a recognition task presented in \cite{scarselli2008graph}, aiming at finding a specific pattern $P$ embedded in a larger graph $G$ of variable sizes with pre-defined communities. The characteristics of this pattern differ from the community in the original graph in terms of size, internal connection probability, and external connection probability. A dataset based on the PATTERN task was later introduced \cite{dwivedi2020benchmarking}, namely SBM-PATTERN. Graphs in SBM-PATTERN are generated using the Stochastic Block Model (SBM), with parameter $p$ being the intra-community linking probability, $p_p$ the intra-pattern linking probability, $q$ the inter-community linking probability, and $q_p$ the community-pattern linking probability. 

The method of generating datasets used in SBM-PATTERN allows us to create new datasets to investigate how the scale of the local substructure impacts the choice of focal length ($FL$) in FFGT, for the following reasons:
(i) In order to ascertain the specific attribution of the nodes, the model needs to not only comprehend the substructural characteristics of the community but also capture the information from distant nodes. This aligns with the applicable scenario of FFGT. 
(ii) By adjusting $p$ and $q$, it is possible to generate graphs with arbitrary sparsity and control the scale of each community, which is desired for the experiment. 

Here we follow the same routine as in \cite{dwivedi2020benchmarking}, but generate intentionally sparser graphs by setting $q$ to 0.01 and $q_p$ to 0.05. With these two probabilities fixed, we create 4 different datasets with different $p$ and $p_p$, i.e. $p=p_p \in \{0.10,0.12,0.14,0.16\}$. When $p$ increases, it only extends the scale of communities without interfering the connections between communities and the pattern. Thus, the datasets generated with a larger $p$ will have communities of a smaller scale. Under this setting, we can test if the scale of the local sub-structures (communities) has a close relation with the optimal focal length $FL$. 
See Appendix \ref{appendix.sbm} for detailed description and parameter settings.

\subsubsection*{Analyses of Results} We train the Vanilla-FFGT with different focal lengths on each dataset, as shown in Table \ref{table.pattern}. Here we use accuracy as the evaluation metric. The results show that when the community scale is small $(p=0.16)$, the model with $FL=1$ performs the best with a statistical significance. As the community scale extends $(p=0.14,0.12)$, the model with $FL=2$ performs on par with that with $FL=1$. When the scale of the community reaches a certain extent $(p=0.10)$, the model with $FL=2$ outperforms that with $FL=1$ with a statistical significance (comparable with $FL=3$). At the same time, the backbone model without the FFGT mechanism always performs the worst. We can conclude from this trend that there is a significant correlation between the optimal focal length $FL^*$ and the scale of the local community. Together with findings in Empirical datasets, we can now conclude that compared with backbone models, the proposed hybrid attention mechanism is more aware of the intrinsic scale of substructures.

\section{Conclusioins}
We propose a paradigm that integrates full-range attention, which focuses on global information, and focal attention, which specializes in comprehending local substructures, which effectively mitigates the deficiency of the Graph Transformer in acquiring information from locality. The compound attention modules interact with each other and can form a more complete picture of the dataset. Unlike previous efforts that supplement local information using MPNNs, the focal attention module we employ overcomes the information loss introduced by stacking layers of MPNNs and can attain more comprehensive information about local substructures. Experimental results on empirical datasets of different substructural scales demonstrate the effectiveness of our framework, seeing that even with a simple backbone model, it can rival SOTA performance. Strikingly the focal length can even grasp the true scale of the side chains in peptides. Our experiments on controllable synthetic graphs further illustrate that the adaptability of this mechanism to the variations in a range of substructures within the graphs. It is especially notable that given the high flexibility of this paradigm, there are numerous avenues for expansions and variants, include linearization of full-range attention module, refining the design of the focal attention module with specific positional encoding, and employing advanced methods of utilizing edge features.

\printbibliography

\begin{appendices}

\section{}
\label{appendix.sbm}

The original SBM-PATTERN dataset \cite{dwivedi2020benchmarking} considers node-level task of graph pattern recognition proposed in \cite{scarselli2008graph}, which aims at finding a specific graph pattern $P$ embedded in larger graphs $G$ of variable sizes. The graphs are generated with the Stochastic Block Model (SBM), which is widely used to model communities in social networks by modulating the intra- and extra-communities connections, thereby controlling the difficulty of the task. A SBM is a random graph generator which assigns communities to each node as follows: any two vertices are connected with a probability $p$ if they belong to the same community, or they are connected with a probability $q$ if they belong to different communities (the value of $q$ acts as a noise level).

\begin{table}[!ht]
\renewcommand\arraystretch{1.4}
\caption{Statistics of four SBM-PATTERN datasets}
\centering
\begin{tabular}{lccccc}
\hline
\textbf{Probability} & \textbf{\#Graphs} & \textbf{Avg.\#nodes} & \textbf{Avg.\#deg} & \textbf{Avg.\#diameters} \\ \hline
$p=0.16$ & 12000 & 125 & 6.13 & 6.15 \\ 
$p=0.14$ & 12000 & 127 & 5.72 & 6.38 \\ 
$p=0.12$ & 12000 & 129 & 5.34 & 6.60 \\ 
$p=0.10$ & 12000 & 130 & 4.85 & 7.00 \\ \hline
\end{tabular}
\label{table.sbm-datasets}
\end{table}

Following the settings in \cite{dwivedi2020benchmarking}, we generate 4 different datasets with different scales of communities. A typical graph $G$ is generated with 5 communities with the size randomly selected in [5, 35]. The extra-probability $q$ for each community is fixed to 0.01, with intra-probability $p \in \{0.10,0.12,0.14,0.16\}$ varying for the 4 datasets. We randomly generate 100 patterns $P$ composed of 20 nodes with a fixed extra-probability $q_p = 0.05$ (i.e., 5 \% of nodes in $P$ are connected to $G$) and intra-probability $p=p_p$. These 100 patterns are randomly assigned to the dataset, with each graph containing one pattern. The output node labels have value 1 if the node belongs to $P$ and value 0 if it is in $G$. Each dataset has 10,000 trainting, 2,000 validation, 2,000 test graphs. Statistics are shown in Table \ref{table.sbm-datasets}. When $p$ increases, the graphs are sparsified and the scale of local community increases.

\end{appendices}

\end{document}